%% file: main.tex
\pdfoutput=1
\documentclass[11pt]{article}

\usepackage{acl}

\usepackage{times}
\usepackage{latexsym}
\usepackage{multirow}
\usepackage{booktabs}
\usepackage{natbib}
\usepackage{titlesec}
\usepackage[ruled,vlined]{algorithm2e}
\usepackage[utf8]{inputenc}
\usepackage{amssymb,amsmath,caption}
\usepackage[hang,flushmargin]{footmisc}
\usepackage{lipsum}
\usepackage{listings}
\usepackage{extarrows}
\usepackage{subfigure}
\usepackage{xspace}
\usepackage{graphicx}
\usepackage{makecell}
\usepackage{ifthen}
\usepackage{xifthen}
\usepackage{wrapfig}
\usepackage{amsthm}
\usepackage{float}

\usepackage[T1]{fontenc}
\usepackage[utf8]{inputenc}

\usepackage{microtype}
\usepackage{colortbl}
\usepackage{makecell}

\newcommand{\vpara}[1]{\vspace{0.07in}\noindent\textbf{#1}\xspace}

\newcommand{\hide}[1]{} %hide

\newcommand\todo[1]{\textbf{\textcolor{orange}{[TODO]}}}

\title{P-Tuning v2: Prompt Tuning Can Be \\ Comparable to Fine-tuning Universally Across Scales and Tasks}

\author{
  Xiao Liu$^{1,2*}$, 
  Kaixuan Ji$^{1*}$, 
  Yicheng Fu$^{1*}$, 
  Weng Lam Tam$^{1}$,
  Zhengxiao Du$^{1,2}$, \\
  {\bf Zhilin Yang$^{1,3\dagger}$, 
  Jie Tang$^{1,2\dagger}$} \\
  $^1$Tsinghua University, KEG \quad $^2$Beijing Academy of Artificial Intelligence (BAAI) \\
  $^3$Shanghai Qi Zhi Institute \\
  \texttt{\{liuxiao21,jkx19,fyc19\}@mails.tsinghua.edu.cn} \\}

\begin{document}
\maketitle

{
\let\thefootnote\relax\footnotetext{
$^\dagger$ corresponding to: Zhilin Yang (zhiliny@tsinghua.edu.cn) and Jie Tang (jietang@tsinghua.edu.cn) }
\let\thefootnote\relax\footnotetext{
$^*$ indicates equal contribution.}
}

\begin{abstract}
    \input{0_abstract}

\end{abstract}

\input{1_introduction}
\input{2_1_problem}

\input{3_method}
\input{4_experiment}
\input{5_conclusion}

\section*{ACKNOWLEDGEMENT}
We would like to thank the anonymous reviewers for their suggestions and comments. 
Jie Tang is supported by the NSFC for Distinguished Young Scholar (61825602) and NSFC (61836013). 
Kaixuan Ji is supported by Tsinghua University Initiative Scientific Research Program and DCST Student Academic Training Program. 

\bibliography{anthology,custom}
\bibliographystyle{acl_natbib}

% \newpage
% ~
\newpage
\appendix
\input{appendix}

\end{document}

%% file: 0_abstract.tex
% Prompt Tuning (or P-tuning), which tunes language models with continuous prompts, has been proved quite promising for many natural language understanding (NLU) tasks.
% However, current P-tuning methods for NLU assume problems as cloze-style tasks, and thus ignore a large family of challenges that are in the form of sequence labelling.
% Here we present SeqP-tuning, an efficient prompt tuning method for sequence labelling. 
% With backbone pre-trained model parameters frozen, by tuning continuous prompts with only 0.1-3\% amount of backbone parameters, SeqP-tuning can be competitive and even better than fine-tuning on typical sequence labelling benchmarks. 
% SeqP-tuning's advantage becomes significant as pre-trained models' sizes grow to billion-scale.

Prompt tuning, which only tunes continuous prompts with a frozen language model, substantially reduces per-task storage and memory usage at training. However, in the context of NLU, prior work reveals that prompt tuning does not perform well for normal-sized pretrained models. We also find that existing methods of prompt tuning cannot handle hard sequence labeling tasks, indicating a lack of universality. We present a novel empirical finding that properly optimized prompt tuning can be universally effective across a wide range of model scales and NLU tasks. It matches the performance of finetuning while having only 0.1\%-3\% tuned parameters. Our method P-Tuning v2 is an implementation of Deep Prompt Tuning \cite{li2021prefix,qin2021learning} optimized and adapted for NLU. Given the universality and simplicity of P-Tuning v2, we believe it can serve as an alternative to finetuning and a strong baseline for future research.\footnote{Our code and data are released at \url{https://github.com/THUDM/P-tuning-v2}.}

\hide{
Prompt tuning, which only tunes continuous prompts with a frozen language model, substantially reduces per-task storage and memory usage at training. However, in the context of NLU, prior work reveals that prompt tuning does not perform well for normal-sized pretrained models. We also find that existing methods of prompt tuning cannot handle hard sequence labeling tasks, indicating a lack of universality. We present a novel empirical finding that properly optimized prompt tuning can be universally effective across a wide range of model scales and NLU tasks. It matches the performance of finetuning while having only 0.1\%-3\% tuned parameters. Our method P-Tuning v2 is not a new method, but a version of prefix-tuning \cite{li2021prefix} optimized and adapted for NLU. Given the universality and simplicity of P-Tuning v2, we believe it can serve as an alternative to finetuning and a strong baseline for future research.\footnote{Our code and data are released at \url{https://github.com/THUDM/P-tuning-v2}.}
}

% To address this issue, we present a universal prompt tuning method P-Tuning v2 that can match the performance of full-model finetuning across a wide range of model scales and NLU tasks, with only 0.1\%-3\% tuned parameters per task. P-Tuning v2 is not a new method but an adaptation of prefix-tuning \cite{li2021prefix} to NLU tasks.

%% file: 1_introduction.tex
\section{Introduction}

Pretrained language models~\cite{radford2019language,devlin2018bert,yang2019xlnet,raffel2019exploring} improve performance on a wide range of natural language understanding (NLU) tasks.
% such as question answering~\cite{2016RajpurkarSQuAD} and textual entailment~\cite{RTE2005}.
A widely-used method, \textbf{fine-tuning}, updates the entire set of model parameters for a target task. While fine-tuning obtains good performance, it is memory-consuming during training because gradients and optimizer states for all parameters must be stored. Moreover, keeping a copy of model parameters for each task during inference is inconvenient since pre-trained models are usually large.

\textbf{Prompting}, on the other hand, freezes all parameters of a pre-trained model and uses a natural language prompt to query a language model \cite{brown2020language}. For example, for sentiment analysis, we can concatenate a sample (e.g., "Amazing movie!") with a prompt ``This movie is [MASK]'' and ask the pre-trained language model to predict the probabilities of masked token being ``good'' and ``bad'' to decide the sample's label. Prompting requires no training at all and stores one single copy of model parameters. However, discrete prompting~\cite{shin2020autoprompt,gao2020making} can lead to suboptimal performance in many cases compared to fine-tuning.

\begin{figure}[t]
    \centering
    \includegraphics[width=\linewidth]{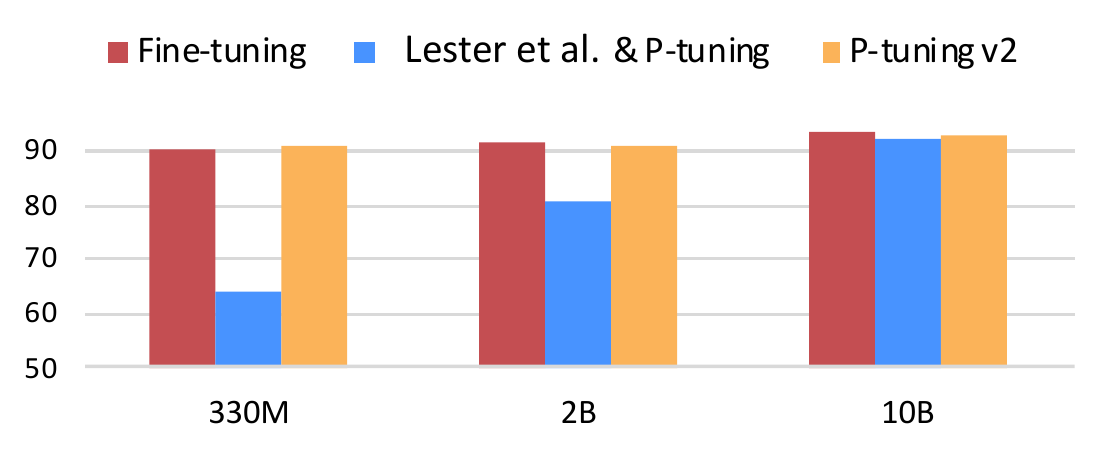}
    \caption{Average scores on RTE, BoolQ and CB of SuperGLUE dev. With 0.1\% task-specific parameters, P-tuning v2 can match fine-tuning across wide scales of pre-trained models, while \citet{lester2021power} \& P-tuning can make it conditionally at 10B scale.}
    \label{fig:example}
    \vspace{-5mm}
\end{figure}

\textbf{Prompt tuning}\footnote{We use ``prompt tuning'' to refer to a class of methods rather than a particular method.} is an idea of tuning only the continuous prompts. Specifically, \citet{liu2021gpt,lester2021power} proposed to add trainable continuous embeddings (also called continuous prompts) to the original sequence of input word embeddings.
% These continuous embeddings (also called continuous prompts) are analogous to discrete manually designed prompts in prompting.
Only the continuous prompts are updated during training. While prompt tuning improves over prompting on many tasks \cite{liu2021gpt,lester2021power,zhong2021factual}, it still underperforms fine-tuning when the model size is not large, specifically less than 10 billion parameters \cite{lester2021power}. Moreover, as shown in our experiments, prompt tuning performs poorly compared to fine-tuning on several hard sequence labeling tasks such as extractive question answering (Cf. Section~\ref{sec:tasks}).

Our main contribution in this paper is a novel empirical finding that properly optimized prompt tuning can be comparable to fine-tuning universally across various model scales and NLU tasks. In contrast to observations in prior work, our discovery reveals the universality and potential of prompt tuning for NLU.

Technically, our approach P-tuning v2 is not conceptually novel. It can be viewed as an optimized and adapted implementation of \textbf{Deep Prompt Tuning}~\cite{li2021prefix,qin2021learning} designed for generation and knowledge probing. The most significant improvement originates from appling continuous prompts for every layer of the pretrained model, instead of the mere input layer. Deep prompt tuning increases the capacity of continuous prompts and closes the gap to fine-tuning across various settings, especially for small models and hard tasks. Moreover, we present a series of critical details of optimization and implementation to ensure finetuning-comparable performance.

Experimental results show that P-tuning v2 matches the performance of fine-tuning at different model scales ranging from 300M to 10B parameters and on various hard sequence tagging tasks such as extractive question answering and named entity recognition. P-tuning v2 has 0.1\% to 3\% trainable parameters per task compared to fine-tuning, which substantially reduces training time memory cost and per-task storage cost.

%% file: 2_1_problem.tex
\section{Preliminaries}

\vpara{NLU Tasks.}
In this work, we categorize NLU challenges into two families: \textit{simple classification tasks} and \textit{hard sequence labeling tasks}.\footnote{Note that the notions of ``simple'' and ``hard'' are specific to prompt tuning, because we find sequence labeling tasks are more challenging for prompt tuning.} Simple classification tasks involve classification over a label space. Most datasets from GLUE~\cite{wang2018glue} and SuperGLUE~\cite{SuperGLUE2019} are in this category. Hard sequence labeling tasks involve classification over a sequence of tokens, such as named entity recognition and extractive question answering.

\vpara{Prompt Tuning.}
Let $\mathcal{V}$ be the vocabulary of a language model $\mathcal{M}$ and let $\mathbf{e}$ be the embedding layer of $\mathcal{M}$. 
% As shown in Figure~\ref{fig:framework}, 
%To classify a film review $\mathbf{x}=$"Amazing movie!" as positive or negative, we can append a prompt "It is [MASK]" to the review and generate the probability of labels ``good'' and ``bad''. 
In the case of discrete prompting~\cite{schick2020s}, prompt tokens \{"It", "is", "[MASK]"\} $\subset \mathcal{V}$ can be used to classify a movie review. For example, given the input text $\mathbf{x}=$"Amazing movie!", the input embedding sequence is formulated as
$
    [\mathbf{e}(\mathbf{x}), \mathbf{e}(\text{"It"}), \mathbf{e}(\text{"is"}), \mathbf{e}(\text{"[MASK]"})]
$.

\citet{lester2021power} and \citet{liu2021gpt} introduce trainable continuous prompts as a substitution to natural language prompts for NLU with the parameters of pretrained language models frozen. Given the trainable continuous embeddings $[h_0,...,h_i]$, the input embedding sequence is written as
$
    [\mathbf{e}(\mathbf{x}), h_0, ..., h_i, \mathbf{e}(\text{"[MASK]"})]
$, as illustrated in Figure \ref{fig:framework}.
Prompt tuning has been proved to be comparable to fine-tuning on 10-billion-parameter models on simple classification tasks~\cite{lester2021power,kim2021changes,liu2021gpt}.

% However, considering the model $\mathcal{M}$ is intrinsically continuous, P-tuning instead proposes to replace prompt tokens with trainable continuous embeddings $[h_0,...,h_i]$ and turn the input sequence into
% $
%     [\mathbf{e}(\mathbf{x}), h_0, ..., h_i, \mathbf{e}(\text{"[MASK]"})]
% $
% (Cf. Figure~\ref{fig:framework} (a)). 

%\xiao{Describe the setting when prompt tuning and P-tuning v2 is using. Describe in details what does simple NLU tasks and hard sequence tagging tasks refer to.}

%% file: 3_method.tex
\begin{figure*}[t]
    \centering
    \includegraphics[width=\linewidth]{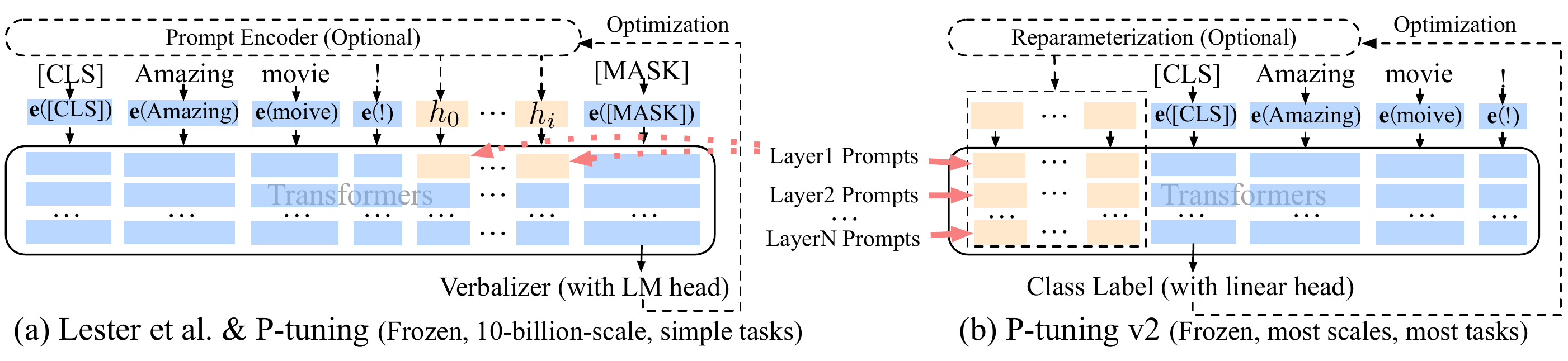}
    \vspace{-7mm}
    \caption{From \citet{lester2021power} \& P-tuning to P-tuning v2. Orange blocks (i.e., $h_0,..., h_i$) refer to trainable prompt embeddings; blue blocks are embeddings stored or computed by frozen pre-trained language models. 
    %Compared to \citet{lester2021power}, P-tuning v2 adds trainable continuous prompts to inputs of every transformer layer independently (as prefix-tuning~\cite{li2021prefix} does). Additionally, P-tuning v2 removes verbalizers with LM head and returns to the traditional class labels with ordinary linear head to allow its task-universality.
    }
    \label{fig:framework}
\end{figure*}

\section{P-Tuning v2}

\subsection{Lack of Universality}
\citet{lester2021power,liu2021gpt} have been proved quite effective in many NLP applications~\cite{wang2021topicrefine,wang2021language,chen2021adaprompt,zheng2021fewnlu,min2021noisy}, but still fall short at replacing fine-tuning due to lack of universality, as discussed below.

% is not yet a qualified alternative to fine-tuning considering following lack of universality.

\vpara{Lack of universality across scales.} \citet{lester2021power} shows that prompt tuning can be comparable to fine-tuning when the model scales to over 10 billion parameters. However, for medium-sized models (from 100M to 1B) that are widely used, prompt tuning performs much worse than fine-tuning.
% there is a significant discrepancy, limiting the applicability of prompt tuning. 

\vpara{Lack of universality across tasks.} Though \citet{lester2021power,liu2021gpt} have shown superiority on some of the NLU benchmarks, the effectiveness of prompt tuning on hard sequence tagging tasks is not verified. 
Sequence tagging predicts a sequence of labels for each input token, which can be harder and incompatible with verbalizers~\cite{schick2020s}. 
In our experiments (Cf. Section~\ref{sec:tasks} and Table~\ref{tab:st}), we show that \citet{lester2021power,liu2021gpt} perform poorly on typical sequence tagging tasks compared to fine-tuning. 

Considering these challenges, we propose P-tuning v2, which adapts deep prompt tuning~\cite{li2021prefix,qin2021learning} as a universal solution across scales and NLU tasks.

%Cite the results from Google to show prompt tuning does not work for normal-sized models
%Cite our results in Section Experiments to show prompt tuning does not work for hard tasks

\subsection{Deep Prompt Tuning}
%\zy{we need a figure to illustrate P-Tuning v2}
%Prefix-tuning~\cite{li2021prefix} was originally proposed for natural language generation (NLG), but we find it very effective for NLU as well. We describe a version of prefix-tuning adapted to NLU. 

In \cite{lester2021power} and \cite{liu2021gpt}, continuous prompts are only inserted into the input embedding sequence (Cf. Figure~\ref{fig:framework} (a)). This leads to two challenges. First, the number of tunable parameters is limited due to the constraints of sequence length. Second, the input embeddings have relatively indirect impact on model predictions.
% Later-layer embeddings at prompt positions are \textbf{computed}, leading to two possible challenges. (1) Limited tunable parameters: most language models currently can only support a maximum sequence length of 512, resulting in a limited prompt length; (2) Limited stability: as the transformer growing deeper, the impact of prompts from the first transformer layer can be quite unexpected due to many intermediate layers' computation, making our optimization not a very smooth one.

To address these challenges, P-tuning v2 employs the idea of deep prompt tuning~\cite{li2021prefix,qin2021learning}.
% does (Cf. Figure~\ref{fig:framework} (b))
% , as a major improvement over P-tuning and \citet{lester2021power}.
As illustrated in Figure \ref{fig:framework}, prompts in different layers are added as prefix tokens. On one hand, P-tuning v2 have more tunable task-specific parameters (from 0.01\% to 0.1\%-3\%) to allow more per-task capacity while being parameter-efficient; on the other hand, prompts added to deeper layers have more direct impact on model predictions (see analysis in Appendix \ref{sec:ablation}).

% (e.g., LayerN Prompts in Figure~\ref{fig:framework}) can have more direct impacts on output predictions (Cf. Appendix~\ref{sec:ablation}).

\subsection{Optimization and Implementation}
There are a few useful details of optimization and implementation for achieving the best performance. 
% notes: %\zy{we need ablation results for the first three, and ablation for multi-layer}

\vpara{Reparameterization.}
Prior works usually leverage a reparameterization encoder such as an MLP \cite{li2021prefix,liu2021gpt} to transform trainable embeddings.
%to increase training speed, robustness, and performance 
% (e.g., MLP for prefix-tuning and LSTM for P-tuning). 
However, for NLU, we discover that its usefulness {depends on tasks and datasets}. For some datasets (e.g., RTE and CoNLL04), MLP brings a consistent improvement; for the others, MLP leads to minimal or even negative effects on the results (e.g., BoolQ and CoNLL12). See Appendix~\ref{sec:ablation} for more analysis.
% show no effect (e.g., BoolQ), sometimes is even worse (e.g., CoNLL12).
%For some datasets, MLP reparameterization brings a consistent improvement over embedding; for others, reparameterization may show no effect, sometimes even worse.

\vpara{Prompt Length.}
The prompt length plays a critical role in P-Tuning v2. We find that different NLU tasks usually achieve their best performance with {different prompt lengths} (Cf. Appendix~\ref{sec:ablation}). Generally, simple classification tasks prefer shorter prompts (less than 20); hard sequence labeling tasks prefer longer ones (around 100).
%, which accords with findings in prefix-tuning~\cite{li2021prefix} where different text generation tasks may have different optimal prompt lengths. See discussions in .

\vpara{Multi-task Learning.}
Multi-task learning jointly optimizes multiple tasks with shared continuous prompts before fine-tuning for individual tasks. Multi-task is optional for P-Tuning v2 but can be used for further {boost performance} by providing a better initialization \cite{gu2021ppt}. 
%On the one hand, the random initialization of continuous prompts brings in difficulties for optimization, which can be alleviated with more training data or task-related unsupervised pre-training~\cite{gu2021ppt}; on the other hand, continuous prompts serve as perfect carriers of task-specific knowledge across tasks and datasets. 
% By providing a better initialization~\cite{gu2021ppt}, it can be a useful complement to P-tuning v2 for some hard sequence labeling tasks
% , denoted as MPT-2 (Cf. Table~\ref{tab:st}).

\vpara{Classification Head.}
Using a language modeling head to predict verbalizers~\cite{schick2020s} has been central for prompt tuning \cite{liu2021gpt}, but we find it unnecessary in a full-data setting and incompatible with sequence labeling. P-tuning v2 instead {applies a randomly-initialized classification head} on top of the tokens as in BERT \cite{2018DevlinBert} (Cf. Figure~\ref{fig:framework}).
% the conventional [CLS] and token classification (Cf. Figure~\ref{fig:framework}) with random-initialized linear heads (Cf. Table~\ref{tab:verbalizer}).
%which turns one-hot class labels into meaningful words to make use of the pre-trained language model head. Despite its potential necessity in a few-shot setting, the verbalizer is not a must in a full-data supervised setting. It hinders the application of prompt tuning to scenarios where we need no-actual-meaning labels and sentence embeddings. Therefore,  See the comparison in Section~\ref{sec:ablation}.

To clarify P-tuning v2's major contribution, we present a conceptual comparison to existing prompt tuning approaches in Table~\ref{tab:comparison}.

\input{tables/comparison}

%% file: tables/comparison.tex
\begin{table}[t]
\footnotesize
\renewcommand\tabcolsep{1.5pt}
\begin{tabular}{@{}lllccc@{}} 
\toprule[1.2pt]
Method                                                      & Task                                  & \makecell[c]{Re-\\param.}     & \makecell[c]{Deep\\PT}    & \makecell[c]{Multi-\\task} & \makecell[c]{No\\verb.} \\ \midrule
\makecell[l]{P-tuning\\\cite{liu2021gpt}}                   & \makecell[l]{KP\\NLU}                 & LSTM                          & -                         & -          & -              \\ \midrule
\makecell[l]{\textsc{PromptTuning}\\\cite{lester2021power}} & NLU                                   & -                             & -                         & \checkmark & -              \\ \midrule
\makecell[l]{Prefix Tuning\\\cite{li2021prefix}}            & NLG                                   & MLP                           & \checkmark                & -          & -              \\ \midrule
\makecell[l]{\textsc{Soft Prompts}\\\cite{qin2021learning}} & KP                                    & -                             & \checkmark                & -          & -              \\ \midrule
\makecell[l]{P-tuning v2\\(Ours)}                           & \makecell[l]{NLU\\SeqTag}             & \it (depends)                 & \checkmark                & \checkmark & \checkmark     \\ 
\bottomrule[1.2pt]
\end{tabular}
\caption{Conceptual comparison between P-tuning v2 and existing Prompt Tuning approaches (KP: Knowledge Probe; SeqTag: Sequence Tagging; Re-param.: Reparameterization; No verb.: No verbalizer).}
\label{tab:comparison}
\end{table}

%% file: 4_experiment.tex
\section{Experiments}

\input{tables/superglue}

\input{tables/seq_tag}

We conduct extensive experiments over different commonly-used pre-trained models and NLU tasks to verify the effectiveness of P-tuning v2. In this work, all methods except for fine-tuning are conducted with \textbf{frozen language model backbones}, which accords with~\cite{lester2021power}'s setting but differs from~\cite{liu2021gpt}'s tuned setting. 
Ratios of task-specific parameters (e.g., 0.1\%) are derived from comparing continuous prompts' parameters with transformers' parameters. 
Another thing to notice is that our experiments are all conducted in the fully-supervised setting rather than few-shot setting.

\vpara{NLU Tasks.}
First, we include datasets from SuperGLUE~\cite{wang2019superglue} to test P-tuning v2's general NLU ability. Additionally, we introduce a suite of sequence labeling tasks, including named entity recognition~\cite{2003sangCoNLL2003,weischedel2013ontonotes,carreras-marquez-2004-introduction}, extractive Question Answering~\cite{2016RajpurkarSQuAD}, and semantic role labeling~\cite{carreras-marquez-2005-introduction,pradhan-etal-2012-conll}).

\vpara{Pre-trained Models.}
We include BERT-large~\cite{2018DevlinBert}, RoBERTa-large~\cite{2019LiuRoberta}, DeBERTa-xlarge~\cite{2020HeDeberta}, GLM-xlarge/xxlarge~\cite{du2021all} for evaluation. They are all bidirectional models designed for NLU tasks, covering a wide range of sizes from about 300M to 10B. 

%\vpara{Comparison Methods.}
%We compare our P-tuning v2 (PT-2) with vanilla fine-tuning (FT), P-tuning \& \citet{lester2021power} (PT). Additionally, for hard tasks regarding the sequence tagging, we present our results on multi-task P-tuning v2 (MPT-2) with more details presented in Section~\ref{sec:tasks}. 

\vpara{Multitask Learning.}
For the multi-task setting, we combine the training sets of the datasets in each task type (e.g., combing all training sets of semantic role labeling). We use separate linear classifiers for each dataset while sharing the continuous prompts (Cf. Appendix~\ref{sec:problem_formulation}).

\subsection{P-tuning v2: Across Scales} \label{sec:scales}
Table~\ref{tab:nlu} presents P-tuning v2's performances across model scales. In SuperGLUE, performances of \citet{lester2021power} and P-tuning at smaller scales can be quite poor. On the contrary, P-tuning v2 matches the fine-tuning performance in all the tasks at a smaller scale. P-tuning v2 even significantly outperforms fine-tuning on RTE.

In terms of larger scales (2B to 10B) with GLM~\cite{du2021all}, the gap between \citet{lester2021power,liu2021gpt} and fine-tuning is gradually narrowed down. On 10B scale, we have a similar observation as \citet{lester2021power} reports, that prompt tuning becomes competitive to fine-tuning. That said, P-tuning v2 is always comparable to fine-tuning at all scales but with only 0.1\% task-specific parameters needed comparing to fine-tuning.

%Additionally, we observe that in some datasets, prompt tuning with RoBERTa-large has poorer performance than BERT-large. Part of the reason is that we empirically find prompt tuning can be quite sensitive to hyper-parameters, and sometimes the tuning just gets trapped. P-tuning v2 can be more stable and robust during tuning. For more details about hyper-parameters, please refer to our code repository.

\subsection{P-tuning v2: Across Tasks} \label{sec:tasks}
%In Section~\ref{sec:scales}, we discuss P-tuning v2's consistent, comparable performance to fine-tuning whatever the scales. However, most tasks on GLUE and SuperGLUE are comparatively simple NLU problems. Another important family of hard NLU challenges lies in sequence tagging, which relates to some more high-level NLP applications, including open information extraction, reading comprehension, and so on. 

%To evaluate P-tuning v2's ability on these hard NLU challenges, we select three typical sequence tagging tasks: Name Entity Recognition, Extractive Question Answering (QA), and Semantic Role Labeling (SRL), altogether eight datasets.  

%\vpara{Results.}
From Table~\ref{tab:st}, we observe that P-tuning v2 can be generally comparable to fine-tuning on all tasks. P-tuning and \citet{lester2021power} show much poorer performance, especially on QA, which might be the most challenging of the three tasks. We also notice that there are some abnormal results of \citet{lester2021power} and P-tuning on SQuAD 2.0. This is probably because SQuAD 2.0 contains unanswerable questions, which causes optimization challenges for single-layer prompt tuning.
% and P-tuning could possibly get a trivial solution. 
Multi-task learning generally brings significant improvements to P-Tuning v2 over most tasks except for QA.
%(which might still be the consequence of mixing all-answerable SQuAD 1.1 and not-answerable SQuAD 2.0), which implies that randomly initialized prompts' potential is under-explored. 

\subsection{Ablation Study}

\input{tables/verbalizer}

\vpara{Verbalizer with LM head v.s. [CLS] label with linear head.}
Verbalizer with LM head has been a central component in previous prompt tuning approaches. However, for P-tuning v2 in a supervised setting, it is affordable to tune a linear head with about several thousand parameters. We present our comparison in Table~\ref{tab:verbalizer}, where we keep other hyper-parameters and only change [CLS] label with linear head to verbalizer with LM head. Here, for simplicity, we use ``true'' and ``false'' for SST-2, RTE and BoolQ; ``true'', ``false'' and ``neutral'' for CB. Results indicate that there is no significant difference between performances of verbalizer and [CLS].

\input{figs/prompt_depth}

\vpara{Prompt depth.}
The main difference between \citet{lester2021power};~\cite{liu2021gpt} and P-tuning v2 is the multi-layer continuous prompts. 
% Intuitively, due to the many non-linear activation functions in intermediate transformer layers, the deeper the transformer layer a prompt locates in, the more direct its impact on the output predictions. 
To verify its exact influence, given a certain number of $k$ layers to add prompts, we select them in both ascending and descending order to add prompts; for the rest layers, we left them untouched.
%change their attention masks for disallowing their prefix prompts to involve in the computation.
As shown in Figure~\ref{fig:depth}, with the same amount of parameters (i.e., num of transformer layers to add prompts), adding them in the descending order is always better than in the ascending order. In the RTE case, only adding prompts to layers 17-24 can yield a very close performance to all layers.
%, further cutting down parameters we may need to tune for matching fine-tuning.

\hide{
\subsection{POS Tagging}

Part-of-speech tagging or POS, to tagging each word with its part of speech in the given sentence, can also be solved as a sequence labeling task. As previous approach has already reached a high accuracy, we only test our model on one dataset.

\vpara{Datasets.}
We adopt LDC95T7 ( WSJ ) to test our model's performance on POS tagging. We use the standard train, develop and test split which use section 0-18 as train set, 19-21 as develop set and 22-24 as test set. The origin dataset formats in Penn-treebank, but when training we only use the POS tags as the only label.

\vpara{Model.}
Given a sentence, the language model gives a hidden representation for each tokens and with these representation, we predict the label of each token. 

}

%% file: tables/superglue.tex
\begin{table*}[t]
    \footnotesize
    \centering
    \renewcommand\tabcolsep{6pt}
    %\ra{1.2}
    \begin{tabular}{@{}
    lp{0.7cm}<{\raggedright}
    p{0.6cm}<{\centering}p{0.6cm}<{\centering}p{0.7cm}<{\centering}p{0.6cm}<{\centering}p{0.6cm}<{\centering}p{0.7cm}<{\centering} 
    p{0.6cm}<{\centering}p{0.6cm}<{\centering}p{0.7cm}<{\centering}p{0.6cm}<{\centering}p{0.6cm}<{\centering}p{0.7cm}
    @{}} \toprule[1.2pt]
                        &\multirow{2}{*}{\#Size}& \multicolumn{3}{c}{BoolQ}  & \multicolumn{3}{c}{CB} & \multicolumn{3}{c}{COPA} & \multicolumn{3}{c}{MultiRC (F1a)} \\ \cmidrule(l){3-5} \cmidrule(l){6-8} \cmidrule(l){9-11} \cmidrule(l){12-14}
                        && FT   & PT    & PT-2    & FT    & PT    & PT-2    & FT    & PT    & PT-2    & FT    & PT    & PT-2  \\
                        \midrule
    BERT$_{\rm large}$   & 335M & \bf 77.7 & 67.2 & \underline{75.8} & \bf 94.6 & 80.4 & \bf 94.6 & \underline{69.0} & 55.0 & \bf 73.0 & \underline{70.5} & 59.6 & \bf 70.6 \\
    RoBERTa$_{\rm large}$& 355M & \bf 86.9 & 62.3 & \underline{84.8}  & \underline{98.2} & 71.4 & \bf 100 & \bf 94.0 & 63.0 & \underline{93.0} & \bf 85.7 & 59.9 & \underline{82.5}\\
    \midrule
    GLM$_{\rm xlarge}$   & 2B   & \bf 88.3 & 79.7 & \underline{87.0} & \bf 96.4 & \underline{76.4} & \bf 96.4 & \bf 93.0 & \underline{92.0} & 91.0 & \underline{84.1} & 77.5 & \bf 84.4 \\
    GLM$_{\rm xxlarge}$  & 10B  & \underline{88.7} & \bf 88.8 & \bf 88.8 & \bf 98.7 & \underline{98.2} & 96.4 & \bf 98.0 & \bf 98.0 & \bf 98.0 & \bf 88.1 & \underline{86.1} & \bf 88.1 \\
    \bottomrule
    \end{tabular}
    \vspace{1mm}
    
    \renewcommand\tabcolsep{6pt}
    \begin{tabular}{@{}
    lp{0.7cm}<{\raggedright}
    p{0.6cm}<{\centering}p{0.6cm}<{\centering}p{0.7cm}<{\centering}p{0.6cm}<{\centering}p{0.6cm}<{\centering}p{0.7cm}<{\centering} 
    p{0.6cm}<{\centering}p{0.6cm}<{\centering}p{0.7cm}<{\centering}p{0.6cm}<{\centering}p{0.6cm}<{\centering}p{0.7cm}
    @{}} \toprule
                        &\multirow{2}{*}{\#Size} & \multicolumn{3}{c}{ReCoRD (F1)} & \multicolumn{3}{c}{RTE} & \multicolumn{3}{c}{WiC} & \multicolumn{3}{c}{WSC}  \\ \cmidrule(l){3-5} \cmidrule(l){6-8} \cmidrule(l){9-11} \cmidrule(l){12-14}
                        && FT   & PT    & PT-2    & FT    & PT    & PT-2    & FT    & PT    & PT-2    & FT    & PT    & PT-2  \\
                        \midrule
    BERT$_{\rm large}$   & 335M & \underline{70.6} & 44.2 & \bf 72.8 & \underline{70.4}    & 53.5  & \bf 78.3  & \underline{74.9} & 63.0 & \bf 75.1 & \bf 68.3 & 64.4 & \bf 68.3 \\
    RoBERTa$_{\rm large}$& 355M & \underline{89.0} & 46.3 & \bf 89.3 & \underline{86.6}    & 58.8  & \bf 89.5  & \bf 75.6 & 56.9 & \underline{73.4} & \underline{63.5} & \bf 64.4 & \underline{63.5}\\ \midrule
    GLM$_{\rm xlarge}$   & 2B   & \underline{91.8} & 82.7 & \bf 91.9 & \bf 90.3 & \underline{85.6} & \bf 90.3 & \bf 74.1 & 71.0 & \underline{72.0} & \bf 95.2 & 87.5 & \underline{92.3} \\
    GLM$_{\rm xxlarge}$  & 10B  & \bf 94.4 & 87.8 & \underline{92.5} & \bf 93.1 & \underline{89.9} & \bf 93.1 & \bf 75.7 & 71.8 & \underline{74.0} & \bf 95.2 & \underline{94.2} & 93.3 \\
    \bottomrule[1.2pt]
    \end{tabular}
    
    \caption{Results on SuperGLUE development set. P-tuning v2 surpasses P-tuning \& \citet{lester2021power} on models smaller than 10B, matching the performance of fine-tuning across different model scales. (FT: fine-tuning; PT: \citet{lester2021power} \& P-tuning; PT-2: P-tuning v2; \textbf{bold}: the best; \underline{underline}: the second best).}
    \vspace{-2mm}
    \label{tab:nlu}
\end{table*}

%% file: tables/seq_tag.tex
\begin{table*}[t]
    \footnotesize
    \centering
    \renewcommand\tabcolsep{4pt}
    %\ra{1.2}
    \begin{tabular}{@{}
    l p{0.7cm}<{\raggedright} 
    p{0.6cm}<{\centering}p{0.7cm}<{\centering}p{0.7cm}<{\centering}p{0.9cm}<{\centering} c 
    p{0.6cm}<{\centering}p{0.7cm}<{\centering}p{0.7cm}<{\centering}p{0.9cm}<{\centering} c 
    p{0.6cm}<{\centering}p{0.7cm}<{\centering}p{0.7cm}<{\centering}p{0.9cm}<{\centering}
    @{}} \toprule[1.2pt]
    
                        &\multirow{2}{*}{\#Size}& \multicolumn{4}{c}{CoNLL03}   & \phantom{} & \multicolumn{4}{c}{OntoNotes 5.0} & \phantom{} & \multicolumn{4}{c}{CoNLL04}   \\
                        \cmidrule{3-6} \cmidrule{8-11} \cmidrule{13-16}
                        && FT    & PT & PT-2 & MPT-2     && FT    & PT & PT-2 & MPT-2    && FT    & PT & PT-2 & MPT-2    \\
                        \midrule
    BERT$_{\rm large}$   & 335M & \bf 92.8   & 81.9 & 90.2  & \underline{91.0} && \bf 89.2  & 74.6 & \underline{86.4}  & 86.3       && \underline{85.6}  & 73.6 & 84.5  & \bf 86.6    \\
    RoBERTa$_{\rm large}$& 355M & \underline{92.6}   & 86.1 & \bf 92.8  & \bf 92.8 && \bf 89.8  & \underline{80.8} & \bf 89.8  & \bf 89.8       && \underline{88.8}  & 76.2 & 88.4  & \bf90.6    \\
    DeBERTa$_{\rm xlarge}$&750M & \bf 93.1   & \underline{90.2} & \bf 93.1  & \bf 93.1 && \underline{90.4}  & 85.1 & \underline{90.4}  & \bf 90.5       && \underline{89.1}  & 82.4 & 86.5  & \bf 90.1    \\
    \bottomrule
    \end{tabular}
    \vspace{1mm}
    
    \renewcommand\tabcolsep{2.6pt}
    %\ra{1.2}
    \begin{tabular}{@{}
    l p{0.7cm}<{\raggedright}
    p{0.6cm}<{\centering}p{0.6cm}<{\centering}p{0.6cm}<{\centering}p{0.6cm}<{\centering}p{0.6cm}<{\centering}p{0.6cm}<{\centering}p{0.6cm}<{\centering}p{0.6cm}<{\centering} c
    p{0.6cm}<{\centering}p{0.6cm}<{\centering}p{0.6cm}<{\centering}p{0.6cm}<{\centering}p{0.6cm}<{\centering}p{0.6cm}<{\centering}p{0.6cm}<{\centering}p{0.6cm}<{\centering}
    @{}} \toprule
                        &\multirow{3}{*}{\#Size}& \multicolumn{8}{c}{SQuAD 1.1 dev (EM / F1)}   & \phantom{} & \multicolumn{8}{c}{SQuAD 2.0 dev (EM / F1)} \\
                        \cmidrule{3-10} \cmidrule{12-19} 
                        && \multicolumn{2}{c}{FT}    & \multicolumn{2}{c}{PT} & \multicolumn{2}{c}{PT-2} & \multicolumn{2}{c}{MPT-2}     && \multicolumn{2}{c}{FT}    & \multicolumn{2}{c}{PT} & \multicolumn{2}{c}{PT-2} & \multicolumn{2}{c}{MPT-2}    \\
                        %&& EM   & F1    & EM    & F1    & EM    & F1    & EM    & F1    && EM   & F1    & EM    & F1    & EM    & F1    & EM    & F1    \\
                        \midrule
    BERT$_{\rm large}$   & 335M & \bf 84.2  & \bf 91.1  & 1.0  & 8.5  & 77.8  & 86.0  & \underline{82.3}  & \underline{89.6}  && \bf 78.7  & \bf 81.9  & 50.2  & 50.2  & 69.7  & 73.5  & \underline{72.7}  & \underline{75.9}  \\
    RoBERTa$_{\rm large}$& 355M & \bf 88.9  & \bf 94.6  & 1.2  & 12.0 & \underline{88.5}  & \underline{94.4}  & 88.0  & 94.1  && \bf 86.5  & \bf 89.4  & 50.2  & 50.2  & 82.1  & 85.5  & \underline{83.4}  & \underline{86.7}  \\
    DeBERTa$_{\rm xlarge}$&750M & \underline{90.1}  & \underline{95.5}  & 2.4 & 19.0 & \bf 90.4  & \bf 95.7  & 89.6  & 95.4  && \underline{88.3}  & \underline{91.1}  & 50.2  & 50.2  & \bf 88.4  & \bf 91.1  & 88.1  & 90.8  \\
    \bottomrule
    \end{tabular}
    \vspace{1mm}
    
    \renewcommand\tabcolsep{4pt}
    %\ra{1.2}
    \begin{tabular}{@{}
    l p{0.7cm}<{\raggedright} 
    p{0.6cm}<{\centering}p{0.7cm}<{\centering}p{0.7cm}<{\centering}p{0.9cm}<{\centering} c 
    p{0.6cm}<{\centering}p{0.7cm}<{\centering}p{0.7cm}<{\centering}p{0.9cm}<{\centering} c 
    p{0.6cm}<{\centering}p{0.7cm}<{\centering}p{0.7cm}<{\centering}p{0.9cm}<{\centering}
    @{}} \toprule
                        &\multirow{2}{*}{\#Size}& \multicolumn{4}{c}{CoNLL12}   & \phantom{} & \multicolumn{4}{c}{CoNLL05 WSJ} & \phantom{} & \multicolumn{4}{c}{CoNLL05 Brown}   \\
                        \cmidrule{3-6} \cmidrule{8-11} \cmidrule{13-16}
                        && FT    & PT & PT-2 & MPT-2     && FT    & PT & PT-2 & MPT-2    && FT    & PT & PT-2 & MPT-2    \\
                        \midrule
    BERT$_{\rm large}$   & 335M & \underline{84.9}   & 64.5. & 83.2 & \bf 85.1 && \bf 88.5  & 76.0  & \underline{86.3}  & \bf 88.5       && \underline{82.7}  & 70.0& 80.7  & \bf 83.1    \\
    RoBERTa$_{\rm large}$& 355M & \bf 86.5   & 67.2  & 84.6  & \underline{86.2} && \bf 90.2  & 76.8  & 89.2  & \underline{90.0}       && \underline{85.6}  & 70.7  & 84.3  & \bf 85.7    \\
    DeBERTa$_{\rm xlarge}$&750M & \underline{86.5}   & 74.1  & 85.7  & \bf 87.1 && \bf 91.2  & 82.3  & \underline{90.6}  & \bf 91.2       && \underline{86.9}  & 77.7  & 86.3  & \bf 87.0    \\
    \bottomrule[1.2pt]
    \end{tabular}
    \caption{Results on Named Entity Recognition (NER), Question Answering (Extractive QA), and Semantic Role Labeling (SRL). All metrics in NER and SRL are micro-f1 score. (FT: fine-tuning; PT: P-tuning \& \citet{lester2021power}; PT-2: P-tuning v2; MPT-2: Multi-task P-tuning v2; \textbf{bold}: the best; \underline{underline}: the second best).}
    % \vspace{-5mm}
    \label{tab:st}
\end{table*}

%% file: tables/verbalizer.tex
\begin{table}[t]
    \footnotesize
    \centering
    \begin{tabular}{p{2.8cm}<{\raggedright}p{0.8cm}<{\centering}p{0.6cm}<{\centering}p{0.6cm}<{\centering}p{0.6cm}<{\centering}} \toprule[1.5pt]
                                & SST-2 & RTE   & BoolQ & CB\\ \midrule
        CLS \& linear head      & 96.3 & 88.4 & 84.8 & 96.4 \\
        Verbalizer \& LM head   & 95.8 & 86.6 & 84.6 & 94.6 \\
        \bottomrule[1.5pt]
    \end{tabular}
    \caption{Comparison between [CLS] label with linear head and verbalizer with LM head on RoBERTa-large.}
    \label{tab:verbalizer}
\end{table}

%% file: figs/prompt_depth.tex
\begin{figure}[t]
    \centering
    \subfigure[RTE]{
        \begin{minipage}[t]{3.5cm}
            \centering
            \includegraphics[width=3.5cm]{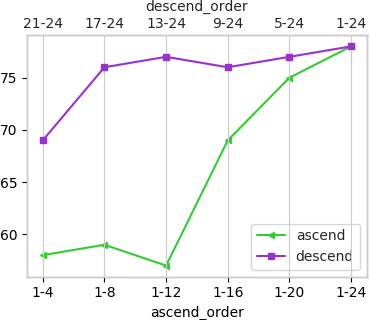}
        \end{minipage}
    }
    \subfigure[BoolQ]{
        \begin{minipage}[t]{3.5cm}
            \centering
            \includegraphics[width=3.5cm]{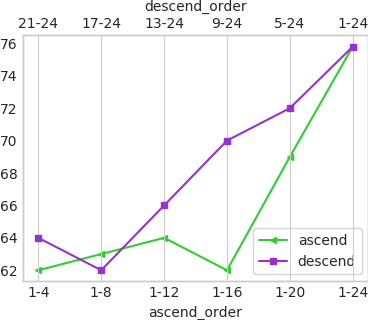}
        \end{minipage}
    }
    \caption{Ablation study on prompt depth using BERT-large. ``[x-y]" refers to the layer-interval we add continuous prompts (e.g., ``21-24'' means we are add prompts to transformer layers from 21 to 24). Same amount of continuous prompts added to deeper transformer layers (i.e., more close to the output layer) can yield a better performance than those added to beginning layers.}
    \label{fig:depth}
        \vspace{-0.2cm}
\end{figure}

%% file: 5_conclusion.tex
\section{Conclusions}
We present P-tuning v2, a prompt tuning method. Despite its relatively limited technical novelty, it contributes to a novel finding that prompt tuning can be comparable to fine-tuning universally across scales (from 330M to 10B parameters) and tasks.
With high accuracy and parameter efficiency, P-Tuning v2 can be a potential alternative for fine-tuning and a strong baseline for future work.
% P-tuning v2 shows consistent improvements on models ranging from 330M to 10B parameters and hard sequence labeling tasks, making it a potential alternative for fine-tuning and a strong baseline for future work.

% comparable to fine-tuning universally across scales and tasks. P-tuning v2 is not conceptually new, but an optimized and adapted prefix-tuning for NLU challenges. P-tuning v2 shows consistent improvements on models ranging from 330M to 10B and hard sequence labeling tasks, making it a potential alternative for fine-tuning and a strong baseline for future work.

%% file: appendix.tex
\input{figs/ablation}

\section{Problem Formulation on Sequence Tagging} \label{sec:problem_formulation}

\vpara{Name entity recognition (NER).} NER aims to predict all spans of words that represent some given classes of entity with a sentence. 
%To show the performance of our model on the NER task, we conduct an experiment on three datasets. 
We adopted CoNLL03~\cite{2003sangCoNLL2003}, OntoNotes 5.0~\cite{weischedel2013ontonotes} and CoNLL04~\cite{carreras-marquez-2004-introduction}. For CoNLL03 and CoNLL04, we trained our model on the standard train-develop-test split. For OntoNotes 5.0, we use the same train, develop, test split as~\cite{xu2021better}. All the datasets are labeled in IOB2 format. We use sequence tagging to solve NER tasks by assigning labels marking the beginning and inside some classes of entity. The language models generate a representation for each token, and we use a linear classifier to predict the labels. We use the official scripts to evaluate the results. For the multi-task setting, we combine the training set of the three datasets for pre-training. We use different linear classifiers for each dataset while sharing the continuous prompts.

\vpara{(Extractive) Question Answering (QA).}
Extractive QA is designed to extract the answer from the context given the context and a question. We use SQuAD~\cite{2016RajpurkarSQuAD} 1.1 and 2.0, in which each answer is within a continuous span of the context. Following tradition, we formulate the problem as sequence tagging by assigning one of the two labels: `start' or `end' to each token and at last selecting the span of the most confident start-end pair as the extracted answer. If the probability of the most confident pair is lower than a threshold, the model will assume the question unanswerable. For the multi-task setting, our training set for pre-training combines the training sets of SQuAD 1.1 and 2.0. When pre-training, we assume that all the questions, regardless of their origin, are possibly unanswerable.

\vpara{Semantic Role Labeling (SRL).}
SRL assigns labels to words or phrases in a sentence that indicate their semantic roles in the sentence. We evaluate P-tuning v2 on CoNLL05~\cite{carreras-marquez-2005-introduction} and CoNLL12~\cite{pradhan-etal-2012-conll}. Since a sentence can have multiple verbs, we add the target verb token to the end of each sentence to help recognize which verb is used for prediction. We classify each word with a linear classifier based on the corresponding semantic role representation. For multi-task setting, the pre-train training set is a combination of the training set of CoNLL05~\cite{carreras-marquez-2005-introduction}, CoNLL12~\cite{pradhan-etal-2012-conll} and propbank-release (a common extend data used for training SRL). The multi-task training strategy is similar to NER.

\section{More Ablation Study} \label{sec:ablation}

Due to the page limit, we present hyper-parameters and architecture designs ablations regarding reparameterization and prompt length in this section. 

\vpara{Embedding v.s. MLP reparameterization.}
In both prefix-tuning~\cite{li2021prefix} and P-tuning~\cite{liu2021gpt}, authors discover the reparameterization to be useful in improving training speed, robustness and performance. However, we conduct experiments to show that the reparameterization effect is inconsistent across different NLU tasks and datasets.

As shown in Figure~\ref{fig:ablation}, in RTE and CoNLL04, MLP reparameterization generally indicates better performance than embedding for almost all prompt lengths. However, in BoolQ, MLP and embedding's results are competitive; in CoNLL12, the embedding consistently outperforms MLP.

\vpara{Prompt Length.}
Prompt length is yet another influential hyper-parameter for P-tuning v2, and its optimal value varies from task to task. From Figure~\ref{fig:ablation}, we observe that for simple NLU tasks, usually, a shorter prompt is enough for the best performance; for hard sequence tasks, usually, a longer prompt than 100 would be helpful.

We also discover that reparameterization has a close bond with optimal prompt length. For example, in RTE, CoNLL04, and BoolQ, MLP reparameterization achieves its optimal result earlier than embedding. This conclusion may contribute some thoughts on P-tuning's optimization properties.

%% file: figs/ablation.tex
\begin{figure*}[t]
    \centering
    \subfigure[NLI: RTE]{
        \begin{minipage}[t]{0.23\linewidth}
            \centering
            \includegraphics[width=\linewidth]{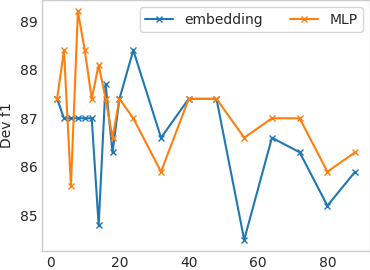}
        \end{minipage}
    }
    \subfigure[NER: CoNLL04]{
        \begin{minipage}[t]{0.23\linewidth}
            \centering
            \includegraphics[width=\linewidth]{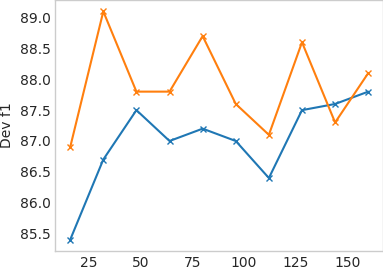}
        \end{minipage}
    }
    \subfigure[MQA: BoolQ]{
        \begin{minipage}[t]{0.23\linewidth}
            \centering
            \includegraphics[width=\linewidth]{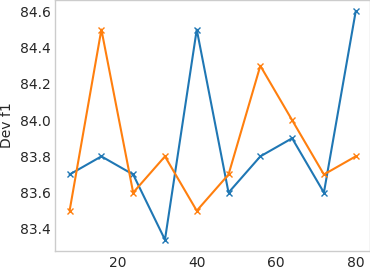}
        \end{minipage}
    }
    \subfigure[SRL: CoNLL12]{
        \begin{minipage}[t]{0.23\linewidth}
            \centering
            \includegraphics[width=\linewidth]{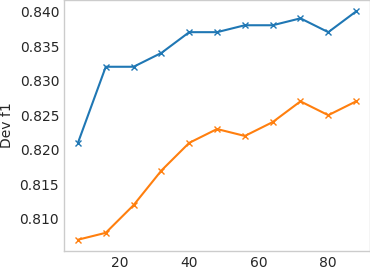}
        \end{minipage}
    }
    \caption{Ablation study on prompt length and reparamerization using RoBERTa-large. The conclusion can be very different given certain NLU task and dataset. (MQA: Multiple-choice QA)}
    \label{fig:ablation}
        \vspace{-0.2cm}
\end{figure*}